\documentclass[runningheads]{llncs}

\usepackage{eccv}

\usepackage{eccvabbrv}

\usepackage{graphicx}
\usepackage{booktabs}

\usepackage[accsupp]{axessibility}  

\usepackage{wrapfig}
\usepackage{multirow}
\usepackage{marvosym}

\usepackage[pagebackref,breaklinks,colorlinks,citecolor=eccvblue]{hyperref}

\usepackage{orcidlink}

\begin{document}

\title{City-Level 3D Surface Reconstruction with Viewpoint Orientation Partitioning and Scene Completion} 

\titlerunning{City-Level 3D Surface Reconstruction}

\author{Liang Han\inst{1} \and
Wenyuan Zhang\inst{1} \and
Junsheng Zhou\inst{1}\textsuperscript{\Letter} \and
Yu-Shen Liu\inst{1}\textsuperscript{\Letter} \and
Zhizhong Han\inst{2}
}

\authorrunning{L.~Han et al.}

\institute{School of Software, Tsinghua University, Beijing, China \\
\email{\{hanl23, zhangwen21, zhou-js24\}@mails.tsinghua.edu.cn, \\liuyushen@tsinghua.edu.cn}
\and
Department of Computer Science, Wayne State University, Detroit, USA
\email{h312h@wayne.edu}
}

\maketitle

\begin{figure}[h!]
  \centering
  \includegraphics[width=\linewidth]{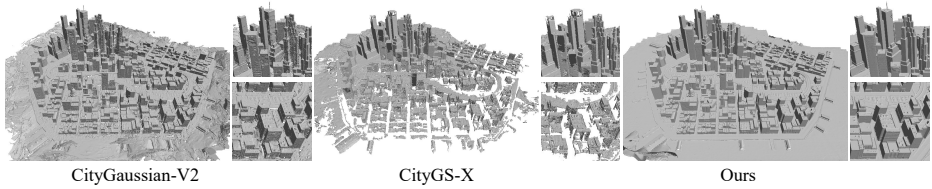}
  \caption{Comparison of reconstruction on the MatrixCity dataset. Compared with state-of-the-art large-scale reconstruction methods CityGaussian-V2 \cite{liu2024citygaussianv2} and CityGS-X \cite{gao2025citygs-x}, Our method achieves more faithful surface reconstruction.}
  \label{fig:teaser}
\end{figure}

\begin{abstract}
Multi-view 3D surface reconstruction is a longstanding challenge in computer vision. 
Although recent large-scale reconstruction methods based on 3D Gaussian Splatting (3DGS) achieve impressive novel-view synthesis, producing high-quality surfaces over large scenes remains difficult, due to complex geometry, long optimization, and limited memory. In this paper, we propose a novel yet simple partitioning method to efficiently and faithfully reconstruct large-scale scene surfaces. Our key insight lies in a scene partitioning method based on viewpoint orientation. This partitioning approach ensures that views with similar orientations are jointly involved for more accurate depth estimations, leading to precise surface reconstructions and balanced computation on multiple GPUs in parallel. 
In addition, we propose a strategy to detect and repair missing regions in the initial point cloud caused by sparse viewpoints or insufficient textures, thereby further improving the geometric quality.
Extensive experiments on the GauU-Scene, MatrixCity, and UrbanScene3D datasets demonstrate that our method outperforms the state-of-the-art approaches in surface reconstruction for large-scale scenes. Project page: \url{https://hanl2010.github.io/VOP-GS}.
\end{abstract}

\section{Introduction}
\label{sec:intro}

3D scene reconstruction is a fundamental task in the field of computer vision. Recently, neural implicit representation methods have made significant progress in this area. In particular, approaches based on Neural Radiance Fields (NeRF) \cite{2021nerf} have demonstrated strong capabilities in novel view synthesis \cite{barron2022mip360, muller2022instantngp} and geometry recovery \cite{wang2021neus, yariv2021volsdf, darmon2022neuralwarp, zhang2025nerfprior, zhang2025monoinstance, zhang2026vrp-udf} using multilayer perceptrons (MLPs) to represent volumetric radiance fields. Meanwhile, 3D Gaussian Splatting (3DGS) \cite{3dgs}, an emerging technique, achieves faster rendering speed than traditional NeRF methods. 3DGS learns a set of Gaussian primitives through differentiable rasterization, thereby improving the efficiency of geometry recovery.

Although 3DGS accelerates radiance field learning through discrete and explicit Gaussian representations, it remains limited by high memory usage and long training time \cite{lin2024vastgaussian, chen2024dogs} in large-scale scene reconstruction.
Currently, most large-scale reconstruction methods \cite{lin2024vastgaussian,wu2025blockgaussian, liu2024citygaussian} adopt a divide-and-conquer strategy with 3DGS, inspired by NeRF-based approaches \cite{blocknerf, wang2024megasurf} that partition the entire scene into spatially adjacent blocks. Each block is represented by a relatively small number of Gaussian primitives and trained with a subset of views, enabling efficient and parallel training on standard multi-GPU setups, which significantly improve the reconstruction quality and speed.

However, spatial partitioning strategies for scene decomposition face challenges such as occlusion, imbalance of complexity, and multiple block fusion. 
The visibility issue arises when a view intersects the boundary of a block. Since Gaussians outside the block are absent during training, regions near the boundary suffer from incomplete visibility, leading to artifacts in reconstruction.
Although existing methods \cite{lin2024vastgaussian, wu2025blockgaussian, liu2024citygaussian} have addressed some of these issues to varying degrees, most of them are primarily designed for novel view synthesis rather than accurate surface reconstruction. Due to inconsistencies and discontinuities at the boundaries between scene partitions, geometric estimation in these regions tends to be error-prone and unstable. 

In order to reconstruct more accurate large-scale geometric surfaces, we aim to design an effective scene partitioning method that primarily relies on camera orientations, and also camera positions serving as a secondary cue. Our key insight is that, under the same optimization constraints, the accuracy of geometric reconstruction is largely affected by the overlap of viewpoints. In general, views with a large overlap are more beneficial for accurate geometry reconstruction, whereas views with little overlap or large angular differences are detrimental. This observation has been justified by many surface reconstruction methods \cite{yao2018mvsnet,na2024uforecon, huang2023neusurf, fatesgs, sparserecon}, which demonstrate that selecting source views with small viewpoint differences and high overlap leads to higher-quality geometry.

Based on this observation, we propose a novel yet simple scene partitioning algorithm that performs lightweight clustering of all input views according to their camera orientations and positions. Our method produces an optimal partition of the scene into viewpoint-consistent groups, each of which exhibits relatively small geometry changes and thereby significantly improves our robustness in surface reconstruction. To further enhance local coherence, we introduce both geometric and photometric constraints among views within each camera group, which effectively preserves scene consistency and recovers fine-grained details at both local and global scales. Each group is then trained separately on multiple GPUs, enabling efficient and distributed learning on large scenes.

\begin{figure}[t]
    \centering
    \includegraphics[width=0.9\linewidth]{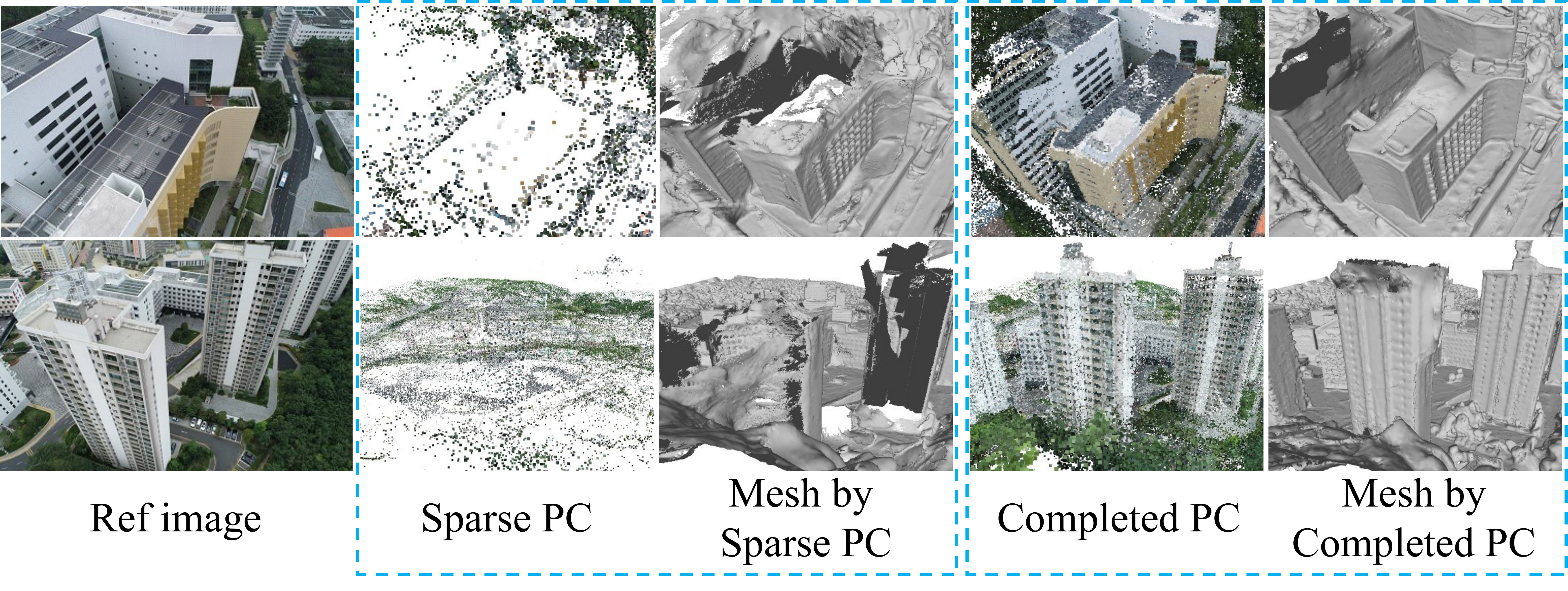}
    \caption{Visual comparison of the point clouds (PC) and corresponding meshes before and after completion. The left column shows the reference image, the middle part presents the COLMAP point cloud and its corresponding mesh reconstruction, and the right part presents the completed point cloud and the corresponding reconstructed mesh.}
    \label{fig:pcd_complete}
\end{figure}

Nevertheless, it is often difficult to ensure 
sufficient image coverage for every region during large-scale data capture. 
Moreover, regions with limited textures make it difficult for Structure-from-Motion (SfM) \cite{schonberger2016sfm} to reconstruct reliable geometry, leading to extremely sparse or even missing 3D points. Consequently, 3DGS-based reconstruction methods often suffer from poor reconstruction quality or incomplete surfaces in these regions. As shown in Fig.~\ref{fig:pcd_complete}, the sparse point cloud predicted by COLMAP \cite{schoenberger2016colmap} exhibits noticeable holes with insufficient image coverage, which further causes defects on surfaces. 
To address this issue, we propose a scene completion strategy to detect and complete the holes in the initial point cloud. Specifically, we first identify the images corresponding to the regions with holes by analyzing the uniformity of the projected points on each image. Then, for each candidate image, we locate its neighboring views and construct image pairs, which are fed into a pretrained image feature matching network \cite{ren2025minima} to predict dense point clouds. Finally, these dense point clouds are used to refine and complete the sparse point cloud, providing a more reliable geometric initialization for large-scale surface reconstruction.

In summary, our main contributions are as follows:
\begin{itemize}
    \item We propose a novel, simple, and effective scene-partitioning strategy that clusters views based on their camera orientations and positions. By grouping highly overlapping views with minimal geometric variation, our method substantially improves the robustness and quality of large-scale surface reconstruction.
    \item We introduce a scene completion method for detecting and completing missing regions in the initial scene point cloud, which further enhances surface reconstruction quality in regions with insufficient view coverage.
    \item Extensive experiments on widely used large-scale scene datasets demonstrate that our method achieves state-of-the-art performance compared with existing large-scene surface reconstruction approaches.
\end{itemize}

\section{Related Work}
\subsection{Object-Centric and Small-Scale Modeling}
Neural implicit representations can be constructed from both images \cite{2021nerf, barron2022mip360, muller2022instantngp, yariv2021volsdf, wang2021neus} and point clouds \cite{li2025learning-normals, li2025pff-net, zhang2025arnet, li2026joint_upsampling_cleaning, shao2025ds-mae, yang2025swin3d++, xu2024geometry_coding, wen2022pmp, xiang2022snowflake}, providing a unified paradigm for 3D scene modeling. 
Neural Radiance Fields (NeRF) \cite{2021nerf} have revolutionized novel view synthesis and implicit 3D reconstruction by leaning a continuous volumetric radiance field from multi-view images. Many subsequent methods \cite{barron2022mip360, muller2022instantngp, wang2021neus, yariv2021volsdf} have improved NeRF’s efficiency, quality, or geometry extraction. Recent advances in 3D Gaussian Splatting (3DGS) \cite{3dgs} enable efficient rendering with explicit, differentiable Gaussian primitives, achieving high-fidelity results in compact scenes. Several extensions \cite{guedon2023sugar, huang20242dgs, lyu20243dgsr, yu2024gof, dai2024gs_surfel, pgsr, li2025gaussianudf, li2026va-gs} enhance surface fidelity by incorporating signed distance functions, normal priors, or geometry-aware regularization, yet remain constrained to relatively small scenes. However, most of these methods focus on object-centric or small-scale scenes, where dense multi-view coverage and limited scene complexity make optimization tractable.

\subsection{Large-Scale Novel View Synthesis}
To scale the neural rendering to large outdoor or urban environments, a series of NeRF-based methods have been proposed. Block-NeRF \cite{blocknerf} pioneered splitting urban street scenes into multiple fixed blocks to enable scalable large scene rendering. However, since each sub-model is trained independently, inconsistencies may arise in geometry and appearance across block boundaries. MegaNeRF \cite{turki2022meganerf} instead leverages visibility-based geometric clustering to assign pixels to region specific sub-NeRF modules. GF-NeRF \cite{shao2025gfnerf} further trains a global coarse model and then fine-tunes local encoders under global guidance to improve boundary consistency and geometry details. BirdNeRF \cite{zhang2024birdnerf} focuses on aerial imagery by partitioning scenes based on camera positions and applying a projection-guided fusion strategy on top of the sub-model training. Meanwhile, 3DGS-based methods such as PyGS \cite{wang2024pygs} and Octree-GS \cite{ren2024octree-gs} adopt level-of-detail (LoD) hierarchies to reduce redundant Gaussians. Both VastGaussian \cite{lin2024vastgaussian} and HUG \cite{su2025hug} introduce visibility-aware block partitioning strategy. BlockGaussian \cite{wu2025blockgaussian} further refines block division based on structural complexity to balance the computational cost of each block. 
OccluGaussian \cite{liu2025occlugaussian} proposes an occlusion-aware scene division strategy for large-scale indoor scene reconstruction to improve efficiency and resource allocation.
Other methods \cite{liu2025efficientgs, feng2025flashgs} improve large-scale scene rendering efficiency by optimizing the Gaussian representation or enhancing the rasterization pipeline. 
DoGaussian \cite{chen2024dogs} performs distributed training on each scene block after partitioning.
Momentum-GS \cite{fan2025momentumgs} introduces scene momentum self-distillation to address the scalability issue of Gaussian decoders in large-scale scene reconstruction.
REUrbanGS \cite{yuan2025reurbangs} designs a controllable LOD generation strategy that enables dynamic LOD selection and real-time rendering under limited computational resources.
Although these methods have achieved high-quality view synthesis in city-scale scenes, they generally focus on appearance modeling rather than precise geometry reconstruction.

\subsection{Large-Scale Surface Reconstruction}
Compared with novel view synthesis, large-scale surface reconstruction has received considerably less attention in recent years. Some methods \cite{wang2024megasurf, yang2025scalable} explicitly target city-scale scene geometric recovery using divide-and-conquer representations based on Signed Distance Fields (SDF), but these approaches typically come with substantial computational overhead. Within the 3DGS family, GigaGS \cite{chen2025gigags} partitions the scene into multiple blocks and imposes multi-view photometric and geometric consistency constraints under an LoD framework. CityGaussian-V2 \cite{liu2024citygaussianv2} extends CityGaussian \cite{liu2024citygaussian} by introducing decomposed-gradient densification and depth regression to mitigate blurring artifacts. CityGS-X \cite{gao2025citygs-x} further pushes efficiency by enabling voxel-level parallelism and eliminates the need for merge-and-partition operations. Despite these advances, most existing methods rely on spatial partitioning strategies and require monocular depth priors for regularization, while insufficiently exploiting cross-view consistency. As a result, achieving robust and efficient large-scale surface reconstruction remains a challenging problem.

\begin{figure*}[t]
    \centering
    \includegraphics[width=1.0\textwidth]{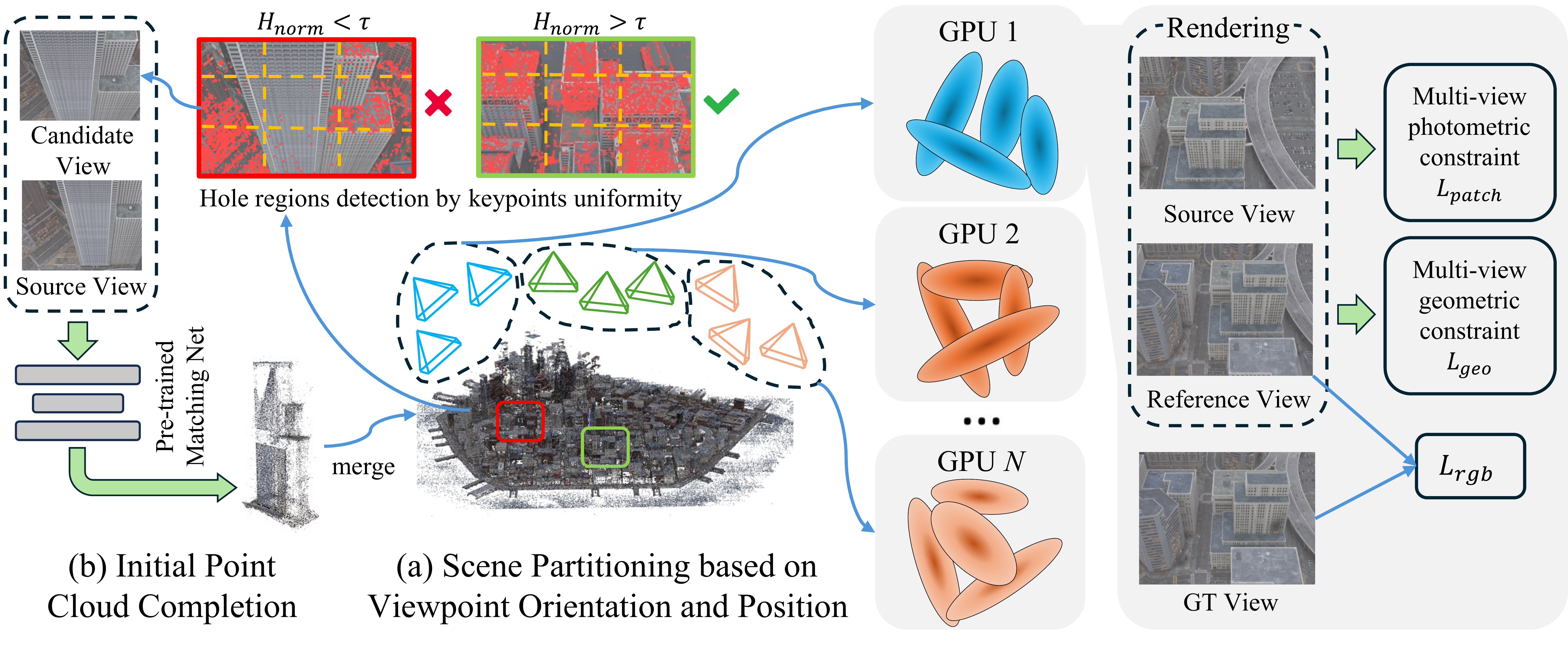}
    \caption{The overview of our method. (a) We group all input views based on their camera orientations and positions, and we train 3DGS with multi-view consistency constraints in each group. (b) We detect regions with holes in the point cloud by analyzing the uniformity of valid keypoints, where the red points in the image indicate the valid keypoints. The corresponding images are then fed into a pretrained matching network to generate dense point clouds, which are used to refine and complete the initialization point cloud of the scene. }
    \label{fig:pipe}
\end{figure*}

\section{Method}
\label{sec:method}

\noindent\textbf{Overview. }The overview of our method is shown in Fig.~\ref{fig:pipe}. We first cluster all input views based on their camera orientations and positions to form multiple training view groups, producing an optimal partition of the scene which exhibits minimal geometric inconsistency. We then analyze the density and uniformity of the projected points from the initial point cloud on each image to identify images covering holes. For these detected images, we employ a pretrained image feature matching network to predict geometric-complete point clouds, which are then used to inpaint the missing areas in the initial scene point cloud.
To further improve the geometric accuracy, for each view we identify a set of source views within the same view group and impose multi-view consistency constraints to enforce local coherence.

\subsection{Scene Partitioning}
\noindent\textbf{Views to Clusters. }To effectively organize a large amount of camera poses in a large-scale scene, we adopt a two-step clustering strategy. The first step clusters cameras based on their orientations, so that cameras with similar viewing directions are grouped together. The cosine distance is used to measure angular similarity,
\begin{equation}
    dist(\mathbf{d}_i, \mathbf{d}_j) = 1-\mathbf{d}_i^\top \mathbf{d}_j,
\end{equation}

\noindent where $\mathbf{d}_i$ and $\mathbf{d}_j$ denote the normalized viewing direction vectors of the $i$-th and $j$-th cameras. DBSCAN \cite{ester1996dbscan} is applied with a cosine distance threshold $\epsilon_{dir}$ to obtain clusters of cameras with similar orientations.

\noindent\textbf{Cluster Splitting into Sub-clusters. }After the initial clustering, some groups may contain much more cameras than other groups. To balance the number of cameras in each group for keeping each cluster at a manageable size, we further subdivide any cluster that contains more than $M_{max}$ cameras.
For each one of such clusters, we estimate the required number of sub-clusters as $K=\lceil \frac{n}{M_{max}} \rceil$, where $n$ donates the number of cameras in this cluster. We then apply K-means clustering to split the cluster into $K$ sub-cluster according to 3D positions $\{ \mathbf{p}_i \}_{i=1}^{n}$. This ensures that each sub-cluster contains no more than $M_{max}$ cameras.

This two-step clustering approach produces robust, viewpoint-consistent camera groups with balanced sizes, enabling efficient parallel processing and local reconstruction in large-scale scenes.

\noindent\textbf{Sub-cluster Refinement. }After splitting each cluster into subsets $\{ G_1, G_2, \dots, \\ G_K \}$ based on camera orientations and positions, we further refine the clustering to ensure that all views within each group have sufficient overlap for reliable geometry optimization, especially for views located near the boundaries of the group. Specifically, for each view $v \in G_k$, we identify a set of source views $S(v) \subseteq \mathcal{V}$ that are favorable for geometry reconstruction of $v$, where $\mathcal{V}$ denotes the set of all views. These source views are then merged into the corresponding cluster, forming an expanded group,
\begin{equation}
    \widetilde{G}_k = G_K \cup \bigcup_{ v \in G_k}^{} S(v).
\end{equation}
\noindent This refinement ensures that the boundary views within each group are supported by sufficient neighboring views for applying local multi-view geometric constraints, while preserving the main structure of the initial clustering.

\begin{figure}[t]
    \centering
    \includegraphics[width=0.9\linewidth]{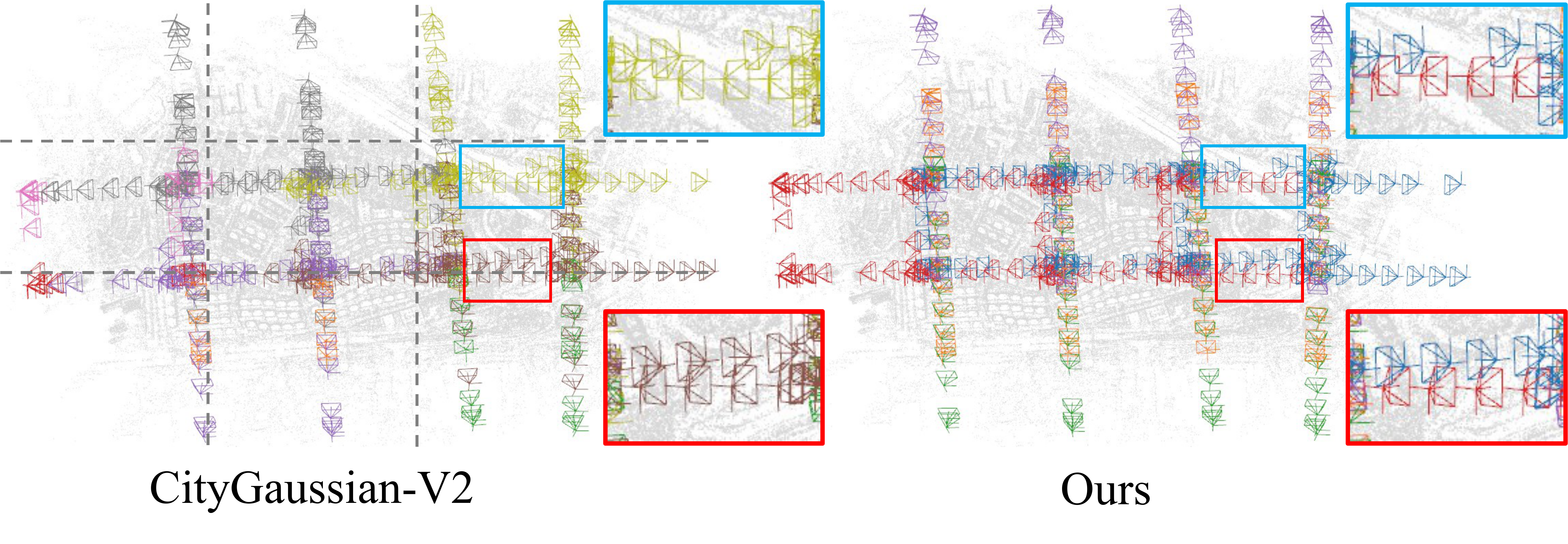}
    \caption{Visual comparison of scene partitioning in our method and CityGaussian-V2. The dashed lines in the left figure indicate the nine spatial partitions used in CityGaussian-V2. Cameras belonging to the same group are shown in the same color. The difference is highlighted.}
    \label{fig:group_compare}
\end{figure}

Fig.~\ref{fig:group_compare} shows a visual comparison of scene partitioning between our method and CityGaussian-V2 \cite{liu2024citygaussianv2}. CityGaussian-V2 merely performs spatial partitioning based on the 3D positions of the scene, which may group cameras with completely opposite viewing directions into the same block. This design can significantly hinder accurate geometry inference in local regions. In contrast, our method clusters views based on camera orientation, ensuring that cameras beneficial to each other's reconstruction are grouped together.

\subsection{Initial Point Cloud Completion}
\noindent\textbf{Motivation. }Incomplete initial point clouds harm the accuracy and completeness of surface reconstruction. While the splitting and cloning strategies in 3DGS can generate new Gaussians to under-reconstructed regions, these mechanisms are largely ineffective when the initial point cloud is extremely sparse or missing entirely. Previous studies \cite{xu2024mvpgs, han2024binocular, guedon2025matcha} have demonstrated that a dense initialization significantly benefits 3DGS-based reconstruction, typically using dense point clouds generated by multi-view stereo (MVS) methods \cite{yao2018mvsnet, schoenberger2016colmap} or pretrained models such as MASt3R \cite{leroy2024mast3r}. However, generating dense point clouds with MVS for large-scale image sets is highly time-consuming, and the pretrained reconstruction models are generally limited to small-scale scenes.

We observe that, in large-scale reconstructions, missing regions in the initial point cloud are typically confined to areas with insufficient viewpoint coverage or poor texture. This insight motivates a targeted approach. Rather than generating a globally dense initialization, we instead focus on detecting missing regions in the initial point cloud and selectively filling them. In this way, we efficiently improve the initialization for 3DGS reconstruction while avoiding the substantial computational cost of full dense point cloud generation.

\noindent\textbf{Hole Detection. }To detect potential missing regions in the initial sparse point cloud, we evaluate the spatial distribution uniformity of valid 3D points projected onto each input image. Intuitively, when the reconstructed 3D points corresponding to a view are densely and uniformly distributed, the regions of the scene covered by that view are likely to be well reconstructed. In contrast, if the projected points are highly clustered or entirely absent in certain regions, the corresponding regions of the scene are likely to contain holes due to insufficient geometric initialization.

In practice, we directly utilize the 2D valid keypoints recorded by COLMAP to detect holes in the point cloud, as these 2D keypoints correspond to the reconstructed 3D points.
Specifically, we divide the image plane into a fixed $N \times N$ grid, and count the number of valid keypoints falling into each cell. The distribution uniformity of valid points is then quantified by the the Shannon entropy,

\begin{equation}
    E = -\sum_{j=1}^{N^2}p_j\log(p_j+\epsilon),
\end{equation}
\begin{equation}
    E_{norm} = \frac{H}{\log(N^2)},
\end{equation}

\noindent where $p_j$ denotes the proportion of points in the $j$-th grid cell, $\epsilon$ is a small constant for numerical stability, and $E_{norm}$ is the normalized entropy.
A lower entropy $E_{norm}$ indicates a more uneven distribution of valid points, suggesting that the corresponding image likely observes a region with insufficient geometric coverage. We therefore regard images with low entropy ($E_{norm} < \tau$) as candidates likely to correspond to missing or sparse regions in the initial point cloud, and select them for subsequent refinement.

\noindent\textbf{Hole Completion. }For each candidate image, its associated neighbor source views are used to construct image pairs, which are fed into a pretrained feature matching network \cite{ren2025minima} to predict dense correspondences. The matched points are triangulated to generate dense 3D points, which are then merged with the original COLMAP sparse point cloud through simple concatenation after the geometric consistency filtering and duplicate removal, producing a refined initialization that fills the detected holes.

\subsection{Training Loss}
Our loss function consists of the multi-view photometric consistency loss $L_{patch}$ and multi-view geometric consistency loss $L_{geo}$ from PGSR \cite{pgsr}, as well as the color reconstruction loss $L_{rgb}$ and depth-normal consistency loss $L_{normal}$ commonly used in 3DGS-based surface reconstruction methods \cite{huang20242dgs, pgsr, yu2024gof}. The final loss function is defined as follows,

\begin{equation}
    L = L_{rgb} + \lambda_{1}L_{patch} + \lambda_{2}L_{geo} + \lambda_{3}L_{normal}.
    \label{train_loss}
\end{equation}

The normalized cross-correlation (NCC) \cite{yoo2009ncc} between patches in the reference view and the source views is used as the multi-view photometric consistency loss, defined as follows,
\begin{equation}
    L_{patch} = \frac{1}{M} \sum (1-NCC(P_r, P_s)),
\end{equation}
where $P_s = H_{rs} P_r$, $P_r$ is the patch on the reference view, and $P_s$ is the corresponding patch on the source view, obtained by warping $P_r$ using the homography matrix $H_{rs}$.

\begin{figure}[t]
    \centering
    \includegraphics[width=1.0\textwidth]{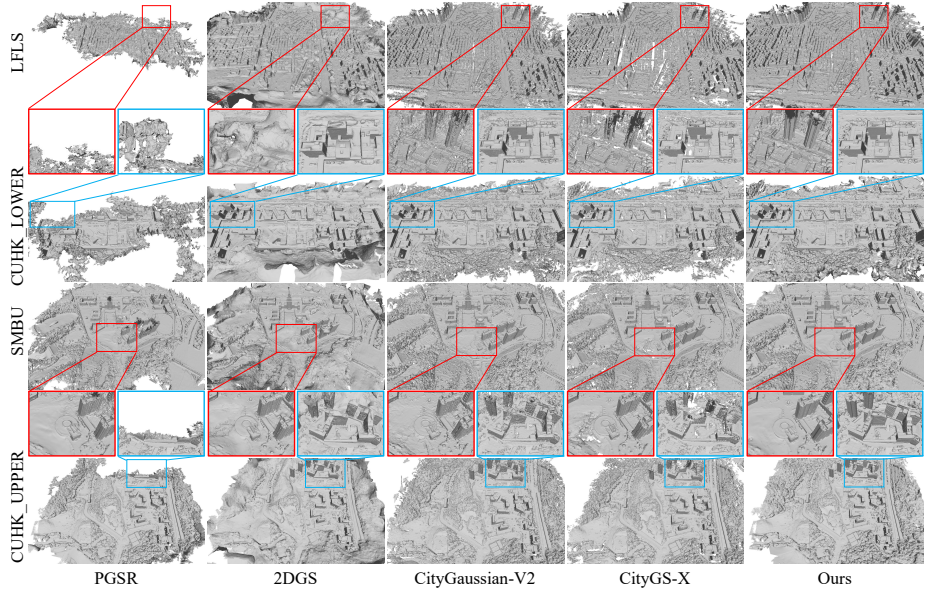}
    \caption{Visual comparison of reconstruction on the GauU-Scene dataset.}
    \label{fig:Gau_compare}
\end{figure}

\section{Experiments}
\subsection{Datasets}
We conduct experiments on three datasets: UrbanScene3D \cite{urbanscene3d}, GauU-Scene \cite{xiong2024gauuscene}, and a small city region from MatrixCity \cite{li2023matrixcity}. Specifically, we use the \textit{Residence} and \textit{Sci-Art} scenes from UrbanScene3D, six scenes from GauU-Scene, and aerial-view images from MatrixCity. Each scene in UrbanScene3D contains more than 2,500 images, while each scene in GauU-Scene includes approximately 400 to 1,200 images. MatrixCity provides over 5,000 images. The spatial extent of these scenes ranges from 0.016 km$^2$ to 2.7 km$^2$.
For preprocessing, we downsample the images from UrbanScene3D and GauU-Scene by a factor of 4. For MatrixCity, we follow the CityGaussian \cite{liu2024citygaussian} and resize the longer edge of each image to 1,600 pixels. 
The sparse point clouds generated by SfM for MatrixCity and UrbanScene are provided by CityGaussian-V2 \cite{liu2024citygaussianv2}.
In addition, GauU-Scene and MatrixCity supply ground-truth point clouds for geometric evaluation.

\subsection{Implementation Details}
We implement our method based on the PGSR \cite{pgsr} framework. Each group is trained for 30K iterations. The threshold $\epsilon_{dir}$ for camera clustering is set to 0.71, which corresponds to an angular threshold of 45 degrees, and the maximum number of cameras in each cluster $M_{max}$ is set to 200. 
We select 4 source views for each reference view. The grid size $N$ for entropy computation is set to 10, and the entropy threshold $\tau$ is set to 0.7. The hyperparameters in \cref{train_loss} are set as $\lambda_{1}$ = 0.15, $\lambda_{2}$ = 0.03, and $\lambda_3$ = 0.05. All experiments are conducted on 8 $\times$ RTX 3090 GPUs.

\subsection{Baselines and Metrics}
We compare our method with state-of-the-art large-scale surface reconstruction approaches based on 3DGS, including CityGaussian-V2 \cite{liu2024citygaussianv2} and CityGS-X \cite{gao2025citygs-x}. 
To compare with the PGSR \cite{pgsr} and 2DGS \cite{huang20242dgs}, we set the training iterations to 100K and reduce the spherical harmonic (SH) degree to 1 to avoid out-of-memory. To evaluate the surface quality, we report the F1 score. We additionally provide the metrics PSNR, SSIM, and LPIPS for novel view rendering in the supplementary materials.

\begin{table*}[t]
\centering
\caption{Quantitative results of F1 score on GauU-Scene dataset. The best results are in \textit{bold}, the second best are \textit{underlined}.}
\resizebox{\linewidth}{!}{
\begin{tabular}{lccccccc}
\toprule
Methods         & CUHK\_LOWER & CUHK\_UPPER & HAV   & LFLS  & SMBU  & SZIIT & Mean F1 $\uparrow$ \\
\midrule
2DGS \cite{huang20242dgs}           & 0.396       & 0.263       & 0.386 & 0.255 & 0.323 & 0.321 & 0.324   \\
PGSR \cite{pgsr}           & 0.102       & 0.493       & \textbf{0.573} & 0.119 & \underline{0.562} & 0.151 & 0.333   \\
CityGaussian-V2 \cite{liu2024citygaussianv2} & 0.491       & 0.481       & \underline{0.570} & \underline{0.574} & 0.425 & 0.495 & 0.506   \\
CityGS-X \cite{gao2025citygs-x}       & \underline{0.508}       & \underline{0.509}       & 0.521 & 0.567 & 0.556 & \underline{0.504} & \underline{0.527}   \\
Ours            & \textbf{0.536}       & \textbf{0.537}       & 0.554 & \textbf{0.594} & \textbf{0.597} & \textbf{0.545} & \textbf{0.560}  \\
\bottomrule
\end{tabular}}
\label{tab:gau_f1}
\end{table*}

\begin{figure*}[]
    \centering
    \includegraphics[width=1.0\textwidth]{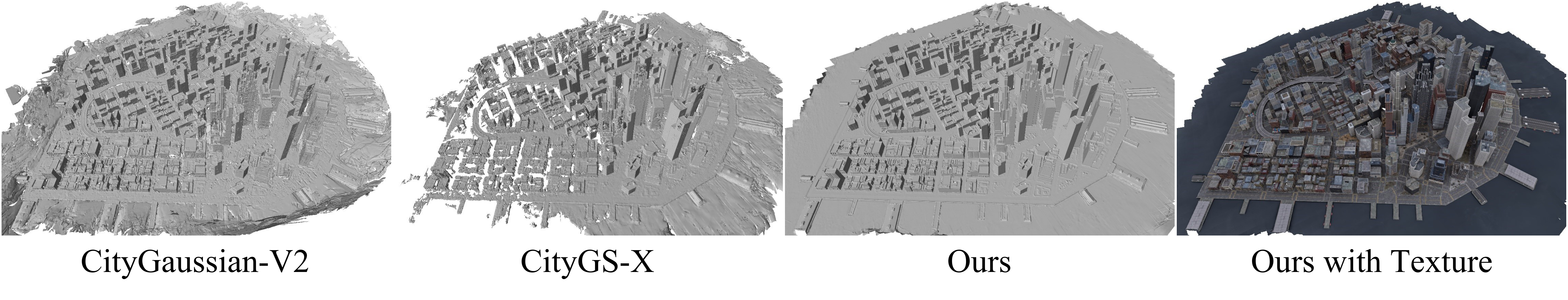}
    \caption{Visual comparison of reconstruction on the MatrixCity dataset.}
    \label{fig:mc_compare}
\end{figure*}

\begin{figure*}[]
    \centering
    \includegraphics[width=1.0\textwidth]{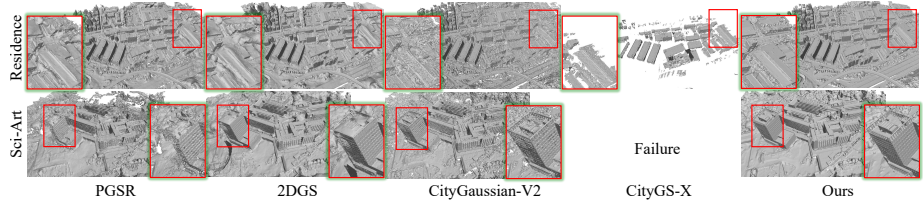}
    \caption{Visual comparison of reconstruction on the UrbanScene3D dataset.}
    \label{fig:urban_compare}
\end{figure*}

\subsection{Quality Comparisons}
\noindent \textbf{GauU-Scene.}
Tab.~\ref{tab:gau_f1} shows the quantitative evaluation results on six scenes from the GauU-Scene \cite{xiong2024gauuscene} dataset. Our method achieves superior overall performance compared with other approaches.
It can be observed that PGSR \cite{pgsr} achieves the best performance on the \textit{HAV} scene, which contains only 424 images and represents a relatively small-scale scene. This indicates that the non-block optimization strategy of PGSR is more advantageous for reconstructing compact scenes.

Fig.~\ref{fig:Gau_compare} provides a visual comparison of surface reconstruction results. 
The reconstruction results of PGSR \cite{pgsr} exhibit severe incompleteness, while 2DGS \cite{huang20242dgs} and CityGaussian-V2 \cite{liu2024citygaussianv2} produces relatively coarse surfaces, and CityGS-X \cite{gao2025citygs-x} also presents noticeable mesh artifacts. In contrast, our method achieves smoother and more complete reconstructions. It can be observed that although CityGaussian-V2 produces more complete surfaces than PGSR on the \textit{SMBU} and \textit{CUHK\_UPPER} scenes, its quantitative results are lower. This is because the ground-truth LiDAR point clouds provided by GauU-Scene \cite{xiong2024gauuscene} contain incomplete edge regions. Following the CityGaussian-V2, we crop out these edge regions and retain only the central regions for quantitative evaluation. Please refer to the supplementary material for more details.

\begin{table}[t]
    \centering
    \caption{Quantitative results of precision, recall and F1 score on MatrixCity dataset. the P and R denote precision and recall respectively. }
    \begin{tabular}{lccc}
    \toprule
    \multirow{2}{*}{Methods} & \multicolumn{3}{c}{MatrixCity} \\
                             & P        & R        & F1       \\
    \midrule
    CityGaussian-V2          & 0.441    & 0.752    & 0.556    \\
    CityGS-X                 & \underline{0.444}    & \underline{0.840}    & \underline{0.581}    \\
    Ours                     & \textbf{0.713}    & \textbf{0.920}    & \textbf{0.803}   \\
    \bottomrule
    \end{tabular}
    \label{tab:matrixcity_f1}
\end{table}

\noindent \textbf{MatrixCity.}
Tab.~\ref{tab:matrixcity_f1} and Fig.~\ref{fig:mc_compare} present the quantitative results and visual comparisons on the MatrixCity \cite{li2023matrixcity} aerial-view dataset. Both PGSR \cite{pgsr} and 2DGS \cite{huang20242dgs} completely fail to reconstruct the geometric surfaces of the MatrixCity due to the large-scale scene and the excessive number of images. CityGaussian-V2 \cite{liu2024citygaussianv2} generates coarse reconstructions, with obvious concavities in dark regions on the buildings. The reconstruction results from CityGS-X \cite{gao2025citygs-x} exhibit significant missing geometry, and do not match those shown in its original paper. Overall, our approach produces the highest-quality reconstructions.

\noindent \textbf{UrbanScene3D.}
Fig.~\ref{fig:urban_compare} illustrates the visual comparison of reconstruction results on the UrbanScene3D \cite{urbanscene3d} dataset. Both PGSR and 2DGS produce distorted and defective surfaces, while CityGaussian-V2 generates coarse reconstructions. CityGS-X fails to reconstruct the \textit{Sci-Art} scene and produces incomplete surfaces in the \textit{Residence} scene. Our method successfully reconstructs both scenes and achieves the best quality. 

\begin{table}[t]
    \centering
    \caption{Efficiency comparison on  GauU-Scene and MatrixCity datasets.}
    \begin{tabular}{lccc}
    \toprule
                    Methods   & CityGaussian-V2 & CityGS-X & Ours \\
    \midrule
    GauU-Scene (hours) & 6.5             & 3        & 2  \\
    MatrixCity (hours) & 13              & 5        & 6.5   \\
    \bottomrule
    \end{tabular}
    \label{tab:efficiency_compare}
\end{table}

\subsection{Efficiency Comparison}
We evaluate the training efficiency of the three methods on the GauU-Scene \cite{xiong2024gauuscene} and MatrixCity \cite{li2023matrixcity} datasets, as shown in Tab.~\ref{tab:efficiency_compare}.
CityGaussian-V2 \cite{liu2024citygaussianv2} adopts a two-stage training strategy and divides each scene into 9 or 16 blocks. These blocks cannot be distributed to 8 GPUs all at once, and each block requires long training time, leading to lower efficiency. CityGS-X \cite{gao2025citygs-x} avoids scene-blocking by employing voxel-level parallelism, which significantly reduces training time. Our method adopts a camera group-based parallel training strategy, where each scene of GauU-Scene contains fewer than 8 groups (up to 7), enabling full GPU utilization in one pass. with each group trained in about 2 hours.
On the MatrixCity dataset, which contains over 5,000 training views, our method also demonstrates competitive training efficiency. 
We cluster all views into 30 groups with the same clustering parameters as other datasets and train over four rounds on 8 GPUs, with a total training time of about 6.5 hours.


\begin{table}[t]
    \centering
    \caption{Ablation studies on the GauU-Scene dataset by clustering the input views using different angular thresholds. The Groups is total number of clusters across all scenes, and the F1 score is the average over all scenes.}
    \begin{tabular}{lccccc}
    \toprule
    Angular Threshold    & 15    & 30    & 45    & 60        \\
    \midrule
    Groups & 36    & 35    & 27    & 27       \\
    F1 Score  & 0.534 & 0.544 & \textbf{0.560} & 0.532  \\
    \bottomrule
    \end{tabular}
    \label{tab:ablation_angle}
\end{table}


\subsection{Ablation Study}
We conduct ablation experiments on the GauU-Scene dataset by clustering the input views using different angular thresholds, while keeping the maximum number of cameras in each cluster no more than $M_{max}=200$. The results are shown in Tab.~\ref{tab:ablation_angle}.
We observe that the best reconstruction quality is achieved when the angular threshold is set to 45 degrees. A threshold that is too small or too large leads to a decline in reconstruction quality. It is worth noting that a smaller angular threshold results in a larger number of clusters, which requires more GPU resources. On the other hand, since we impose a maximum number of cameras per group in the second-stage clustering based on camera positions, increasing the angular threshold does not reduce the number of clusters. However, it introduces greater variation in camera orientations within each group, which in turn degrades the quality of the reconstruction.

\begin{figure}[]
    \centering    
    \includegraphics[width=1.0\linewidth]{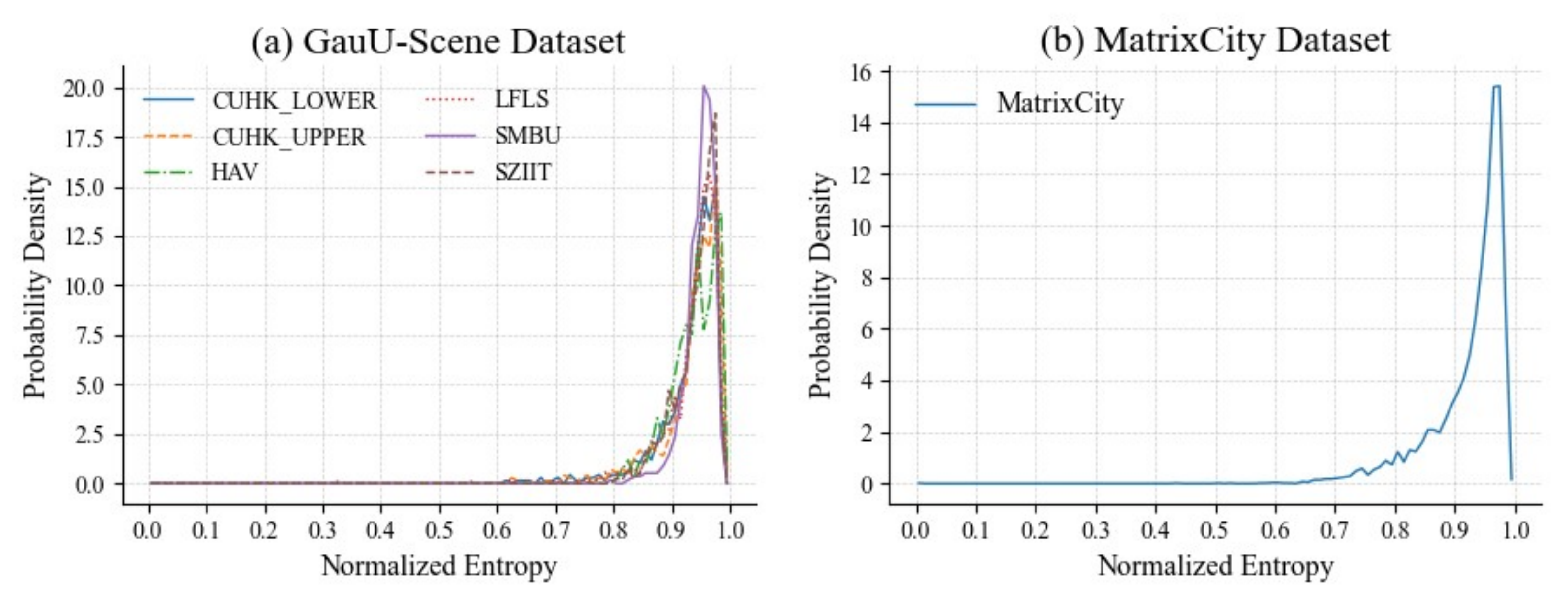}
    \captionof{figure}{Distribution of normalized entropy on the Gau-Uscene and MatrixCity Datasets.}
    \label{fig:entropy_distribution}
\end{figure}

To validate the effectiveness of the initialization point cloud enhancement and determine the appropriate entropy threshold $\tau$, we conduct ablation studies on the GauU-Scene \cite{xiong2024gauuscene} and MatrixCity \cite{li2023matrixcity} datasets. We first compute the probability density distribution of image entropy $E_{norm}$ for all scenes in GauU-Scene and MatrixCity, as shown in Fig.~\ref{fig:entropy_distribution}. It can be observed that the entropy $E_{norm}$ of valid points in images across different scenes exhibits a similar distribution, mostly ranging between 0.6 and 1.0, indicating that a unified entropy threshold 
can be applied across different scenes. 



\begin{table}[t]
    \centering
    \captionof{table}{Ablation of entropy theshold $\tau$ on the GauU-Scene and MatrixCity datasets.}
    \begin{tabular}{lcccccc}
    \toprule
    $\tau$     & w/o   & 0.6   & 0.7   & 0.8   & 0.85  & 0.9   \\
    \midrule
    GauU-Scene & 0.546 & 0.551 & 0.560 & 0.561 & 0.563 & 0.565 \\
    MatrixCity & 0.772 & 0.797 & 0.803 & 0.805 &  0.806 & OOM  \\
    \bottomrule
    \end{tabular}
    \label{tab:ablation_tau}
\end{table}

We then gradually increase the entropy threshold $\tau$ starting from 0.6 and perform reconstruction and quantitative evaluation for scenes in GauU-Scene and MatrixCity, with results summarized in \cref{tab:ablation_tau}. 
The results show that, for both GauU-Scene and MatrixCity, reconstruction quality progressively improves as the entropy threshold $\tau$ increases, i.e., as the number of images used to enhance the initial point cloud grows. More importantly, an entropy threshold of 0.6 for the initial point cloud enhancement leads to a notable improvement in reconstruction performance, even when only a few images are used. However, as the $\tau$ increases, the performance gain becomes marginal, while the point cloud completion process becomes more time-consuming. In addition, the excessive number of points in the initial point cloud may lead to GPU memory exhaustion. Considering both the entropy distribution (as shown in Fig.~\ref{fig:entropy_distribution}) and computational efficiency, we set the entropy threshold to 0.7. It is worth noting that even without scene completion, our method still outperforms other baselines on MatrixCity and achieves comparable performance to CityGS-X on GauU-Scene.

\begin{figure}[t]
    \centering
    \includegraphics[width=1.0\linewidth]{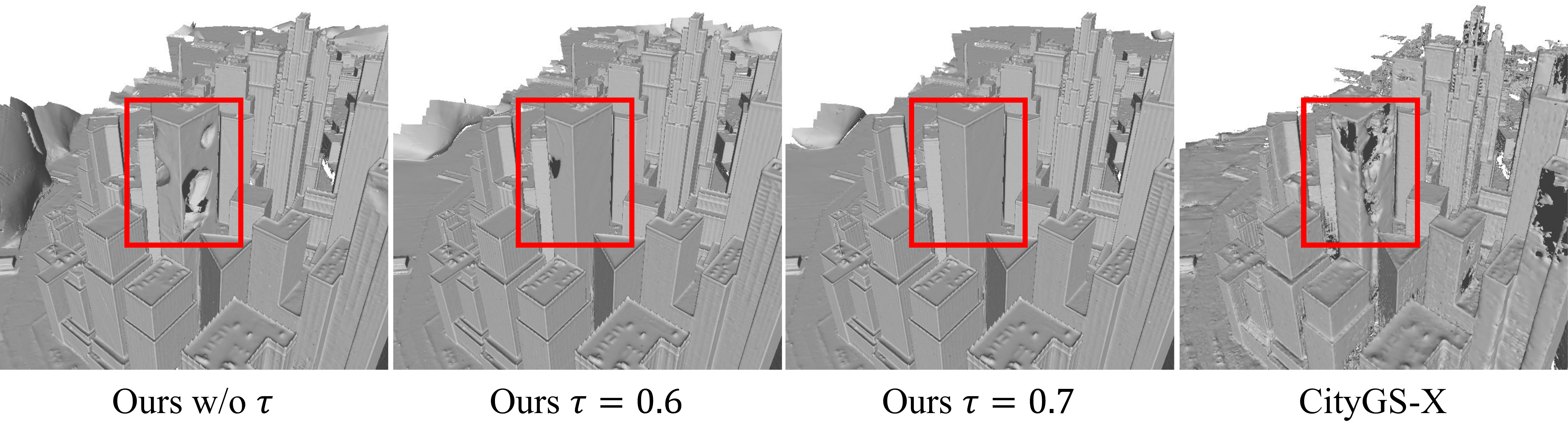}
    \caption{Visualization of reconstruction with varying entropy threshold $\tau$ values.}
    \label{fig:ablation_vis}
\end{figure}

Fig.~\ref{fig:ablation_vis} shows the reconstruction results under different values of $\tau$. It can be observed that as $\tau$ increases, the defects in the mesh gradually decrease.

To further verify the effect of completing the initial point cloud on reconstruction quality of other methods, we use the completed point cloud as initialization for 2DGS \cite{huang20242dgs}, PGSR \cite{pgsr}, CityGaussian-V2 \cite{liu2024citygaussianv2} and CityGS-X \cite{gao2025citygs-x} on the GauU-Scene \cite{xiong2024gauuscene} dataset and conduct quantitative evaluations of the reconstruction results, as shown in Tab.~\ref{tab:init_compare}. The results show that incorporating the completed point clouds leads to performance gains for all methods,
demonstrating that local point cloud completion effectively enhances reconstruction quality. We follow CityGaussian‑V2 \cite{liu2024citygaussianv2} and crop the boundaries of the provided ground-truth point clouds, retaining only the central region for evaluation. Note that the evaluation regions for each scene are therefore subsets of the full scene, while the completion of the initialization point cloud may occur outside these predefined regions, which explains why the improvement is not always noticeable in some cases.

\begin{table}[]
    \centering
    \caption{Quantitative results of F1 score on the GauU-Scene dataset with different initial point clouds (PC).}
    \resizebox{\linewidth}{!}{
    \begin{tabular}{llccccccc}
    \toprule
Methods               & Init.        & CUHK\_LOWER & CUHK\_UPPER & HAV   & LFLS  & SMBU  & SZIIT & Mean F1 $\uparrow$ \\
    \midrule
\multirow{2}{*}{2DGS} & Sparse PC    & 0.396       & 0.263       & 0.386 & 0.255 &                          0.323 & 0.321 & 0.324   \\
                      & Completed PC & 0.412       & 0.312       & 0.387 & 0.254 & 0.325 & 0.324 & \textbf{0.336}   \\
    \midrule
\multirow{2}{*}{PGSR} & Sparse PC    & 0.102       & 0.493       & 0.573 & 0.119 &                          0.562 & 0.151 & 0.333   \\
                      & Completed PC & 0.345       & 0.530       & 0.574 & 0.119 & 0.561 & 0.164 & \textbf{0.382}  \\
    \midrule
\multirow{2}{*}{CityGaussian-V2} & Sparse PC & 0.491 & 0.481     & 0.570 & 0.574 &
                      0.425  & 0.495 & 0.506    \\
                       & Completed PC &  0.520      & 0.493      & 0.569 & 0.579 &
                      0.425  & 0.495 & \textbf{0.514}     \\
    \midrule
\multirow{2}{*}{CityGS-X}& Sparse PC & 0.508       & 0.509       & 0.521 & 0.567 &
                      0.556 & 0.504 & 0.527          \\
                       & Completed PC & 0.531       & 0.518      & 0.521  & 0.568 &
                      0.557 & 0.504 & \textbf{0.533}          \\
    \bottomrule
\end{tabular}}
    \label{tab:init_compare}
\end{table}

\section{Conclusion}
We propose a novel method for large-scale scene reconstruction. We group the input views based on camera orientation and position, ensuring that the views with overlaps
are clustered together to recover more accurate scene geometry. Meanwhile, we detect and complete regions with missing geometry in the initial point cloud, which further enhances surface reconstruction quality in regions with insufficient view coverage. Extensive experiments on the GauU-Scene, MatrixCity, and UrbanScene3D datasets demonstrate that our method outperforms the latest large-scale surface reconstruction methods.

\section*{Acknowledgements}
The corresponding authors are Yu-Shen Liu and Junsheng Zhou. This work was partially supported by Deep Earth Probe and Mineral Resources Exploration -- National Science and Technology Major Project (2024ZD1003405), and the National Natural Science Foundation of China (62272263).

%
%
\bibliographystyle{splncs04}
\bibliography{main}
\end{document}